\documentclass[11pt,a4paper]{article}
\usepackage[hyperref]{naaclhlt2019}
\usepackage{times}
\usepackage{latexsym}
\usepackage{comment}
\usepackage{graphicx}
\usepackage{url}
\usepackage{tabularx}
\usepackage{float}
\usepackage{amsfonts}

\usepackage{multirow}
\usepackage{color, colortbl}
\usepackage{xcolor}
\aclfinalcopy 
\def\aclpaperid{737X-C5G9F6D3P5} 

\DeclareMathAlphabet\mathbfcal{OMS}{cmsy}{b}{n}

\title{Joint Learning of Pre-Trained and Random Units for Domain Adaptation in Part-of-Speech Tagging}

\author{Sara Meftah$^{\ast}$, Youssef Tamaazousti$^{\mp}$, Nasredine Semmar$^{\ast}$, Hassane Essafi$^{\ast}$, Fatiha Sadat$^{+}$ \\
  $^{\ast}$CEA, LIST, LASTI, France\\ 
  $^{\mp}$MIT, CSAIL, USA  \\
  $^{+}$UQ{\`A}M, Montr{\'e}al, Canada  \\
     \{firstname.lastname\}@cea.fr , ytamaaz@mit.edu , sadat.fatiha@uqam.ca  \\}         

\date{}
\begin{document}
\maketitle

\begin{abstract}
Fine-tuning neural networks is widely used to transfer valuable knowledge from high-resource to low-resource domains. 
In a standard fine-tuning scheme, source and target problems are trained using the same architecture. 
Although capable of adapting to new domains, pre-trained units struggle with learning uncommon target-specific patterns. 
In this paper, we propose to augment the target-network with normalised, weighted and randomly initialised units that beget a better adaptation while maintaining the valuable source knowledge. 
Our experiments on POS tagging of social media texts (Tweets domain) demonstrate that our method achieves state-of-the-art performances on 3 commonly used datasets.
\end{abstract}

\section{Introduction}
\label{sec:introduction}
\vspace{-0.2cm}

POS tagging is a sequence labelling problem, that consists on assigning to each sentence' word, its disambiguated POS tag (\textit{e.g.}, Pronoun, Noun) in the phrasal context in which the word is used. 
Such information is useful for higher-level applications, such as 
machine-translation \cite{niehues2017exploiting} or cross-lingual information retrieval~\cite{semmar2008evaluating}.  

\begin{figure}[tb!]
\centerline{\includegraphics[scale=0.33]{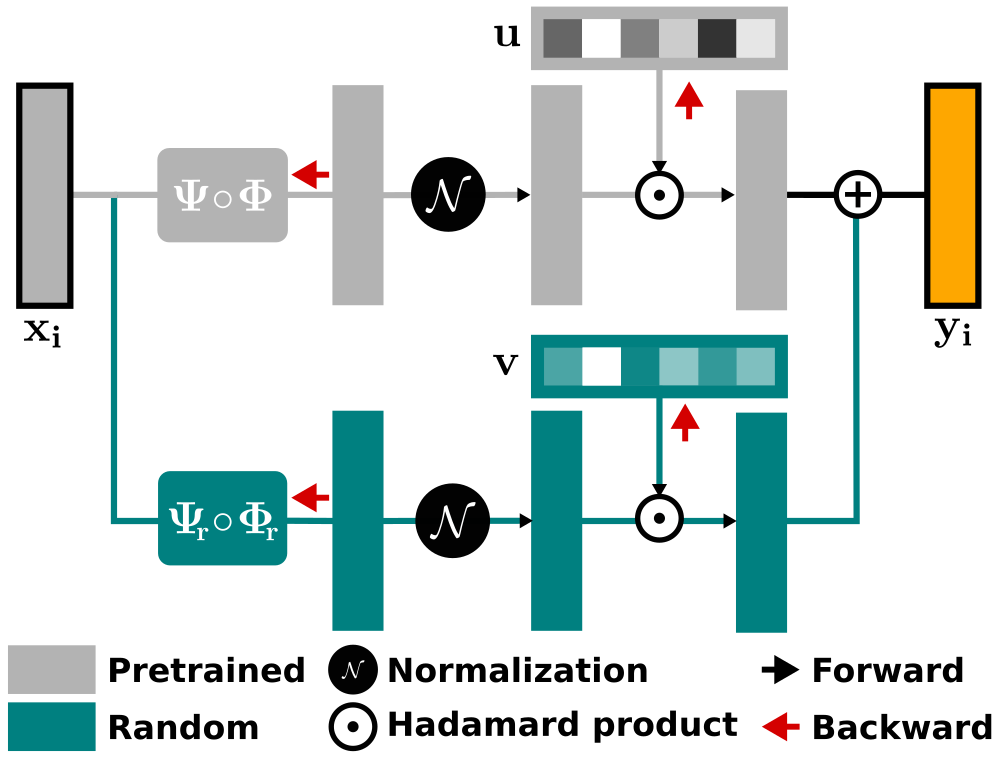}}
\vspace{-0.2cm}
\caption{
Given a word representation $x_i$, a BiLSTM ($\mathbf{\Phi}$) models the sequence, and a FC layer ($\bf\Psi$) performs classification. 
In standard fine-tuning, the units are pre-trained on a large source-dataset then adapted to the target one. 
In this work, we propose to add randomly initialised units (green branch) and jointly adapt them with pre-trained ones (gray branch). 
An element-wise sum is further applied to merge the two branches. 
Before merging, we balance the different behaviours of pre-trained and random units, using an independent normalisation ($\mathbfcal{N}$). 
Finally we let the network learn which of pre-trained or random neurons are more suited for every class, by adding learnable weighting vectors ($u$ and $v$ initialised with 1-values) on the FC layers. 
}
\label{scheme_1}
\vspace{-0.5cm}
\end{figure}


One of the best approaches for POS tagging of social media text~\cite{meftah2018neural}, is transfer-learning, which relies on a neural-network learned on a \textit{source-dataset} with sufficient annotated data, then further adapted to the problem of interest  (\textit{target-dataset}). 
While this approach is known to be very effective \cite{zennaki2019neural}, because it takes benefit from pre-trained neurons, it has one main drawback by design. 
Indeed, it has been shown in computer-vision~\cite{zhou2018interpreting} that, when fine-tuning on scenes a model pre-trained on objects, it is the neuron firing on the \textit{white dog} object that became highly sensitive to the \textit{white waterfall} scene. 
Simply said, pre-trained neurons are \textit{biased} by what they have learned in the source-dataset. 
This is also the case on NLP (see experiments). 
Consequently, pre-trained units struggle with learning patterns specific to the target-dataset (\textit{e.g.}, ``wanna'' or ``gonna'' in the Tweets domain). 
This last is non-desirable, since it has been shown recently~\cite{zhou2018revisiting} that such specific units are important for performance. 
To overcome this drawback, one can propose to take benefit from randomly initialised units, that are by design non-biased. 
However, it is common to face small target-datasets that contain too few data to learn such neurons from scratch. 
Hence, in such setting, it is hard to learn random units that fire on specific patterns and generalise well. 

In this article, we propose a hybrid method that takes benefit from both worlds, without their drawbacks. 
It consists in augmenting the source-network (set of pre-trained units) with randomly initialised units and jointly learn them. 
We call our method \textbf{PretRand} (\textbf{Pret}rained and \textbf{Rand}om units) and illustrate it in Fig.~\ref{scheme_1}. 
The main difficulty is forcing the network to consider random units, because they have different behaviours than pre-trained ones.  
Indeed, while these last strongly fire discriminatively on many words, these first do not fire on any word at the initial stage of fine-tuning. 
Therefore, random units do \textit{not} significantly contribute to the computation of gradients and are thus slowly updated. 
To overcome this problem, we proposed to independently normalise pre-trained and random layers. 
This last balances their range of activations and thus forces the network to consider them, both. 
Last but not least, we do not know which of pre-trained and random units are the best for every class-predictor, thus we propose to learn weighting vectors on top of each branch. 

Evaluation was carried on 3 POS tagging Tweets datasets in a transfer-learning setting. 
Our method outperforms SOTA methods and significantly surpasses fairly comparable baselines. 

\section{Proposed Method: PretRand}
\label{sec:proposed_method} 
\vspace{-0.2cm}

\subsection{Base Model} 
\label{base_model}
\vspace{-0.2cm}
Given an input sentence $S$~$=$~$[w_{1},\ldots,w_{n}]$ of $n$ successive tokens \(w_i\), the goal of a POS tagger is to predict the POS-tag $c_i$~$\in$~$\mathcal{C}$ of every \(w_{i}\), with $\mathcal{C}$~$\in$~$\mathbb{R}^C$ being the tag-set.  
Hence, for our base model, we used a common sequence labelling model 
which first, computes for each token \(w_i\), a word-level embedding (denoted $\mathbf{\Upsilon_w}$) and character-level embedding using biLSTM encoder ($\mathbf{\Upsilon_c}$), and concatenates them to get a final representation $x_i$. 
Second, it feeds the later representation into a biLSTM features extractor (denoted $\mathbf{\Phi}$) that outputs a hidden representation, that is itself fed into a fully-connected (FC) layer (denoted $\mathbf{\Psi}$) for classification. 
Formally, given $w_i$, the logits are obtained using: $\hat{y}_{w_i} = \mathbf{\Psi} \circ \mathbf{\Phi} \circ \mathbf{\Upsilon}(w_i)$, with $\mathbf{\Upsilon}$ being the concatenation of the output of $\mathbf{\Upsilon_c}$ and $\mathbf{\Upsilon_w}$ for $w_i$. 
In a standard fine-tuning scheme \cite{meftah2018using}, $\mathbf{\Upsilon}$ and $\mathbf{\Phi}$ are pre-trained on the \textit{source-task} and $\mathbf{\Psi}$ is randomly initialised. 
Then, the three modules are further jointly trained on the \textit{target-task} by minimising a Softmax Cross-Entropy (SCE) loss using the SGD algorithm. 

\subsection{Adding Random Branch}
\label{method_add_random}
\vspace{-0.2cm}
As mentioned in the introduction, pre-trained neurons are biased by design, thus limited. 
This motivated our proposal to augment the pre-trained branch with additional random units (as illustrated in Fig.~\ref{scheme_1}). 
To do so, theoretically one can add the new units in any layer of the base model. 
However in practice, we have to make a trade-off between performances and the number of parameters (model complexity). 
Thus, given that deep layers are more task-specific than shallow ones~\cite{peters2018deep,mou2016transferable}, and that word embeddings (shallow layers) contain the majority of parameters, we choose to expand only the top layers. 
With this choice, we desirably increase the complexity of the model only by $1.02\times$ compared to the base one. 
In terms of the layers expanded,  
we specifically add $k$ units to $\mathbf{\Phi}$ resulting in an extra biLSTM layer: $\mathbf{\Phi_{r}}$ ($r$ for rand); and $C$ units in $\mathbf{\Psi}$ resulting in an extra FC layer: $\mathbf{\Psi_{r}}$. 
Hence, for every $w_i$, the additional random branch predicts class-probabilities following: $\hat{y}^r_{w_i} = \mathbf{\Psi_r} \circ \mathbf{\Phi_r} (x_i)$ (with $x_i = \mathbf{\Upsilon}(w_i)$). 
Note that, having two FC layers obviously outputs two predictions per class (one from the pre-trained FC \(\hat{y}_{w_{i}}^{p}\) and one from the random  \(\hat{y}_{w_{i}}^{r}\)), that thus need to be merged. 
Hence, to get the final predictions, we simply apply an element-wise sum between the output of both branches: $\hat{y}_{w_{i}} = \hat{y}_{w_{i}}^{p} \oplus \hat{y}_{w_{i}}^{r}$. 
As in the classical scheme, SCE is minimised but here, both branches are trained jointly. 

\subsection{Independent Normalisation}
\label{method_norm}
\begin{figure}[tb!]

\begingroup
\setlength{\tabcolsep}{2pt} 
\renewcommand{\arraystretch}{0.5} 

\begin{tabular}{c c}
\includegraphics[scale=0.21]{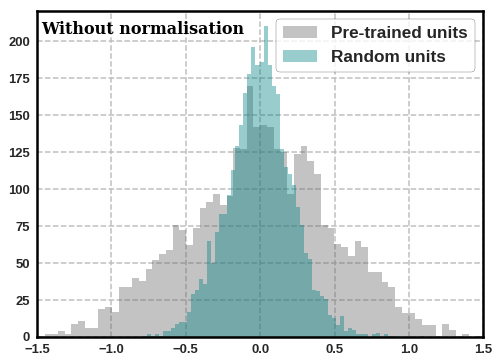}
& \includegraphics[scale=0.21]{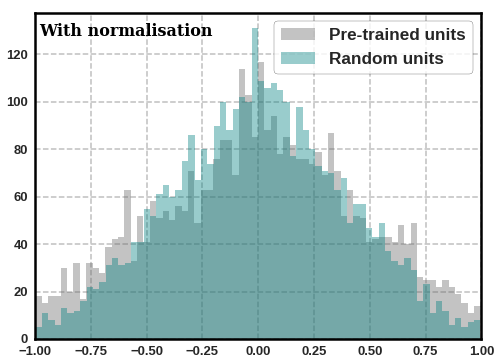}
\end{tabular}

\endgroup

\vspace{-0.3cm}
\caption{Distributions of learned weight-values for the randomly initialised (green) and pre-trained (gray) fully-connected layers after joint training. 
Left: without normalisation, right: with normalisation.
}
\label{FCL_weights}
\vspace{-0.5cm}
\end{figure}

\vspace{-0.2cm}
Nevertheless, while at the initial stage of fine-tuning, the pre-trained units are strongly firing on many words, the random ones are firing very weakly. 
As stated in some computer-vision works~\citep{liu2015parsenet,tamaazousti2018universal}, the later setting causes an absorption of the weights, outputs and thus gradients of the random units by the pre-trained ones, which thus makes them useless at the end. 
We encountered the same problem with textual data on the POS-tagging problem. 
Indeed, as illustrated in the left plot of Fig.\ref{FCL_weights}, at the end of training, the distribution of the random units' weights is still absorbed (closer to zero) by that of the pre-trained ones. 

To prompt the two classifiers to work cooperatively, we normalise (using an \(\ell_p\)-norm) both of them independently before merging them. 
Formally, we apply $\mathbfcal{N}_p(x)$~$=$~$\frac{x}{||x||_p}$ on $\hat{y}_{w_{i}}^{p}$ and $\hat{y}_{w_{i}}^{r}$. 
The normalisation is desirably solving the weights absorption problem since at the end of the training, the distributions of the pre-trained and random weights become very similar (right of Fig.~\ref{FCL_weights}). 

Furthermore, we have observed that despite the normalisation, the performances of the pre-trained classifiers were still much better than the randomly initialised ones. Thus, to make them more competitive, we propose to start with optimising only the randomly initialised units while freezing the pre-trained ones, then, launch the joint training. This is called random++ in the following.

\subsection{Learnable Weighting Vectors}
\label{method_weighted_vectors}
\vspace{-0.15cm}
Back to the extra predictor (FC layer of random branch), it is important to note that both branches are equally important for making a decision for every class, \textit{i.e.,} no weight is applied on the dimensions of $\hat{y}_{w_{i}}^{p}$ and $\hat{y}_{w_{i}}^{r}$. 
However, this latter is sub-optimal since we, a priori, do not know which kind of units (random or pre-trained) is better for making a decision. 
Consequently, we propose to weight the contribution of the predictions for each class. 
For this end, instead of simply performing an element-wise sum between the random and pre-trained predictions, we first \textit{weight} each of them with learnable weighting vectors, then compute a Hadamard product with their associated normalised predictions;  
the learnable vectors $u \in \mathbb{R}^C$ and $v \in \mathbb{R}^C$, respectively corresponding to the pre-trained and random branch, are initialised with 1-values and are learned by SGD. 
Formally, the final predictions are computed following:  \(\hat{y}_{w_{i}} = u \odot \mathbfcal{N}_{p}(\hat{y}_{w_{i}}^{p}) \oplus v \odot \mathbfcal{N}_{p}(\hat{y}_{w_{i}}^{r})\).

\section{Experiments}
\label{sec:experiments}
\vspace{-0.25cm}

\subsection{Implementation Details}
\label{section_experimental_setup}
\vspace{-0.1cm}

In the word-level embeddings, tokens are lower-cased while the character-level component still retains access to the capitalisation information. 
We set the 
character embedding dimension at 50, the dimension of hidden states of the character-level biLSTM at 100 and used 300-dimensional word-level embeddings. 
The latter were pre-loaded from publicly available Glove pre-trained vectors on 42 billions words from a web crawling and containing 1.9M words~\cite{pennington2014glove}. Note that, these embeddings are also updated during fine-tuning. 
For biLSTM (token-level feature extractor), we set the number of units of the pre-trained branch to 200 and experimented our approach with $k$ added random-units, with \(k \in \{50,100,150,200\}\). 
For the normalisation, we used \(\ell_2\)-norm. 
Finally, in all experiments, training was performed using SGD with momentum and mini-batches of 8 sentences. 
Evidently, all the hyperparameters have been cross-validated.  

\subsection{Datasets}
\label{section_datasets}
\vspace{-0.15cm}

For the source-dataset, we used the Penn-Tree-Bank (PTB) of Wall Street Journal (\textbf{WSJ}), a large 
English dataset containing 
1.2M+ tokens from the newswire domain annotated with the PTB tag-set. 
Regarding the target-datasets, we used three datasets in the Tweets domain: 
\textbf{TPoS}~\citep{ritter2011named}, annotated with 40 tags 
; \textbf{ARK}~\cite{owoputi2013improved} 
containing 25 coarse tags; 
and the recent \textbf{TweeBank}~\cite{liu2018parsing} 
containing 17 tags 
(PTB universal tag-set). 
The number of tokens in the datasets are given in Table~\ref{data_sets}.


\begin{table}[t!]
\small
\centering
\begin{small}
 \begin{tabular}{| l | c | c | c |}
 \hline
 \bf Corpus & \bf TPoS & \bf Ark &  \bf TweeBank \\
 \hline\hline

 \bf Train & 10,652 & 26,594 &  24,753 \\
 \bf Dev & 2,242 & n/a &  11,742 \\
 \bf Test & 2,291 & 7,707 & 19,112 \\

 \hline
\end{tabular}
\end{small}
\vspace{-0.2cm}
\caption{Number of tokens in every used dataset.}
\vspace{-0.6cm}
\label{data_sets}
\end{table}

\definecolor{Gray}{gray}{0.75}
\newcolumntype{g}{>{\columncolor{Gray}}c}

\begin{table*}
\small
\centering
 \begin{tabular}{|l | l | c c | c | c   c | g |}
 \hline
 \multirow{ 2}{*}{\bf Method} & \multirow{ 2}{*}{\textbf{\#params}} & \multicolumn{2}{ c|}{\textbf{TPoS}} & \textbf{ArK} & \multicolumn{2}{c|}{\textbf{TweeBank}} &  \\
  \cline{3-7}
  &  & \it Dev & \it Test  & \it Test & \it Dev &  \it Test  & \multirow{-2}{*}{ \textbf{Avg} }\\
 \hline
 \hline
\bf GATE~\citep{derczynski2013twitter} & n/a & 89.37 & 88.69 & n/a & n/a &  n/a & n/a \\
\bf GATE-bootstrap~\citep{derczynski2013twitter}  & n/a & n/a & 90.54 &  n/a & n/a &  n/a & n/a \\
\bf ARK~\citep{owoputi2013improved} & n/a &  n/a & 90.40   & \underline{93.2} & n/a &  \underline{94.6} & n/a \\
\bf  TPANN~\citep{gui2017part} & n/a & \underline{91.08} &  \underline{90.92} & 92.8 & n/a & n/a & n/a\\
\hline
\bf Random-200 & $1\times$ &  88.32 & 87.76 &  90.67 & 91.20 &  91.56 & 89.90 \\
\bf Random-400 & $1.03\times$ & 89.01 & 88.89 & 90.99 & 91.38 &  91.63 & 90.38\\
\bf Standard fine-tuning & $1\times$  &  90.96 & 90.7 &  91.72 & \underline{92.59} & 92.99  & \underline{91.79}\\
\bf Ensemble Model (2 rand)& $2\times$  &  89.20   &  88.8  & 91.36 & 91.73 &  92.05 & 90.62\\
\bf Ensemble Model (1 pret + 1 rand)& $2\times$   &  89.77 & 88.61 & 91.41 & 92.57 &  92.85 & 91.04\\
\hline

\bf PretRand (Ours)& $1.02\times$  & \textbf{91.56}  & \textbf{91.46} & \textbf{93.77}  & \textbf{94.51} & \textbf{94.95} & \textbf{93.24}\\


\hline
\end{tabular}
\vspace{-0.3cm}
\caption{
Comparison of our method to state-of-the-art (top) and baselines (bottom) in terms of token-level accuracy (in \%) on 3 Tweets datasets. 
Note that, baselines are more fairly comparable to our method.  
In the second and last columns, we respectively highlighted the number of parameters and the average performance on the 3 datasets.  
}



\label{perfs_all_data_sets}
\vspace{-0.3cm}
\end{table*}

\subsection{Comparison Methods}
\label{section_sota}
\vspace{-0.15cm}

To assess the POS tagging performances of our PretRand model, we compared it to 5 baselines: \textbf{Random-200} and \textbf{Random-400}: randomly initialised neural model with 200 and 400 biLSTM's units; 
\textbf{Fine-tuning}: pre-trained neural model, fine-tuned with the standard scheme; 
\textbf{Ensemble (2 rand)}: averaging the predictions of two base models randomly initialised and learned independently (with different random initialisation) on Tweets datasets; 
and \textbf{Ensemble (1 pret + 1 rand)}: same as the previous but with one pre-trained on WSJ and the other randomly initialised. 

We also compared it to the 3 best SOTA methods: 
\newcite{derczynski2013twitter} (GATE) is a model based on HMMs with a set of normalisation rules, external dictionaries and lexical features. They experiment it on TPoS, with WSJ and 32K tokens from the NPS IRC corpus. They also used 1.5M additional training tokens annotated by vote-constrained bootstrapping (GATE-bootstrap). 
\newcite{owoputi2013improved} proposed a model based on first-order Maximum Entropy Markov Model (MEMM) with greedy decoding and using brown clustering and careful hand-engineered features.
Recently, \newcite{gui2017part} proposed TPANN that uses adversarial training to leverage huge amounts of unlabelled Tweets.
\vspace{-0.1cm}

\begin{figure}[tb!]
\centerline{\includegraphics[scale=0.32]{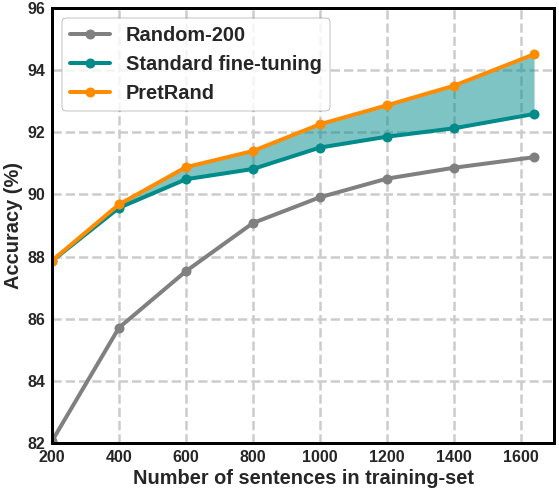}}
\vspace{-0.2cm}
\caption{Performances (on dev-set of TweeBank) according different training-set sizes for the target-dataset. 
Transparent green highlights the difference between our PretRand and standard fine-tuning.
}
\label{accuracy_curve}
\vspace{-0.2cm}
\end{figure}

\begin{table}
\small
\centering
 \begin{tabular}{|l | c c c | g |}
 \hline
 \textbf{Method} & \textbf{TPoS} & \textbf{ArK} & \textbf{TweeBank} & \textbf{Avg} \\
 \hline
 \hline
\bf PretRand & \textbf{91.46}   & \textbf{93.77}  & \textbf{94.95} & \textbf{93.39} \\
\bf -learnVect  &  91.25 &  93.46 &  94.59  &  93.10\\
\bf -random\textsuperscript{++} & 90.97 &  93.11 &  94.13 & 92.73\\
\bf -l2 norm & 90.76 &  92.11 &  93.38   &  92.08\\
\hline
\end{tabular}
\vspace{-0.2cm}
\caption{
\textbf{Ablation study}, Token level accuracy (in \%) when progressively ablating PretRand components. 
}
\label{perfs_jplru_ablation}
\vspace{-0.4cm}
\end{table}

\subsection{Results}
\vspace{-0.15cm}

\begin{figure}[tb!]
\centerline{\includegraphics[scale=0.30]{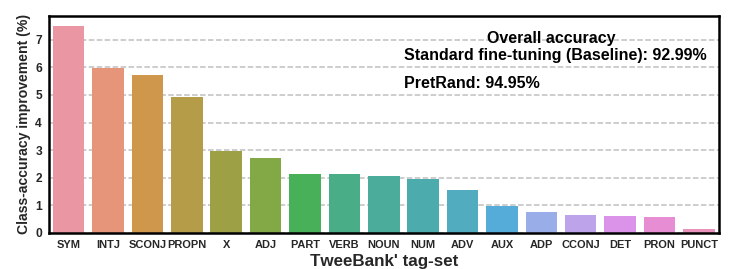}}
\vspace{-0.3cm}
\caption{
Sorted class-accuracy improvement (\%) on TweeBank  of PretRand compared to fine-tuning.
}
\label{acc_class}
\vspace{-0.5cm}
\end{figure}

From the results given in Table~\ref{perfs_all_data_sets}, one can first see that our approach outperforms the SOTA and baseline methods on all the datasets. 
More interestingly, PretRand significantly outperforms the popular fine-tuning baseline by +1.4\% absolute point on average and is better on all classes (see per-class improvement on Fig.~\ref{acc_class}). 
PretRand also outperforms the challenging Ensemble Model by a large margin (+2.2\%), while using much less parameters. 
This clearly highlights the difference of our method with ensemble methods and the importance of having a shared word representation as well as our normalisation and weighting learnable vectors during training. 
A key asset of PretRand, is that it uses only 0.02\% more parameters compared to the fine-tuning baseline. 

An interesting experiment is to evaluate the gain of performance of PretRand compared to fine-tuning, according different target-datasets' sizes. 
From the results in Fig.~\ref{accuracy_curve}, PretRand has desirably a bigger gain with bigger \textit{target-task} datasets, which clearly means that the more target training-data, the more interesting our method will be.

To assess the contribution of different components of PretRand, we performed an ablation study. Specifically, we successively ablated the main components of PretRand, namely, the learnable vectors (learnVect), the longer training for random units (random++) and the normalisation (\(\ell_2\)-norm). 
From the results in Table~\ref{perfs_jplru_ablation}, we can observe that the performances are only marginally better than standard fine-tuning when ablating the three components from PretRand. 
More importantly, adding each of them successively, makes the performances significantly better, which highlights the importance of every component.

\section{Analysis}
\label{sec:analysis}
\vspace{-0.3cm}


\begin{figure}[tb!]
\centerline{\includegraphics[scale=0.20]{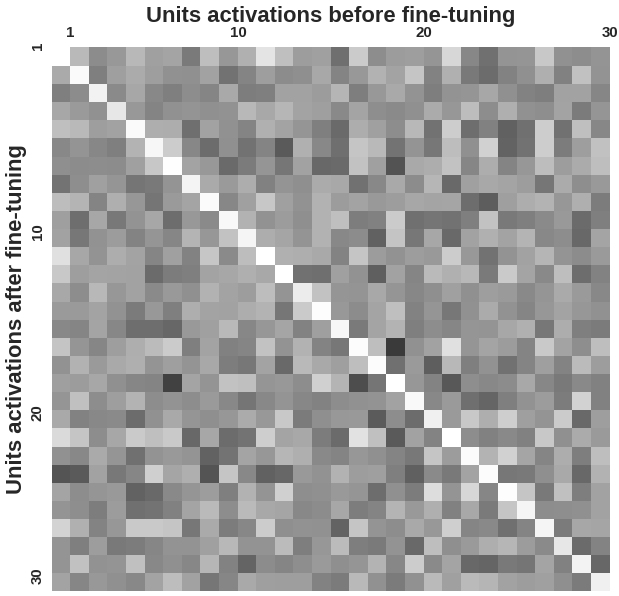}}
\vspace{-0.3cm}
\caption{
Correlation between units' activations before fine-tuning (columns) and after fine-tuning (rows).  
}
\label{units_corr}
\vspace{-0.5cm}
\end{figure}

\textbf{Bias when fine-tuning pre-trained units}\\
\noindent
Here our goal is to highlight that as in~\cite{zhou2018interpreting}, pre-trained units can be biased in the standard fine-tuning scheme. 
To do so, we follow~\cite{tamaazousti2017mucale} and analyse the units of $\mathbf{\Phi}$ (biLSTM layer) \textit{before} 
and \textit{after} fine-tuning.  
Specifically, we compute the Pearson's correlation between all the units of the layer before and after fine-tuning. 
Here, a unit is represented by the random variable being the concatenation of its output activations from all the validation samples of the TweeBank dataset. From the resulting correlation matrix illustrated in Fig.~\ref{units_corr}, one can clearly observe the white diagonal, highlighting the fact that, every unit after fine-tuning is more correlated with itself before fine-tuning than with any other unit. 
This clearly confirms our initial motivation that pre-trained units are highly biased to what they have learned in the source-dataset, making them limited to learn patterns specific to the target-dataset. 

\begin{figure}[tb!]
\centerline{\includegraphics[scale=0.26]{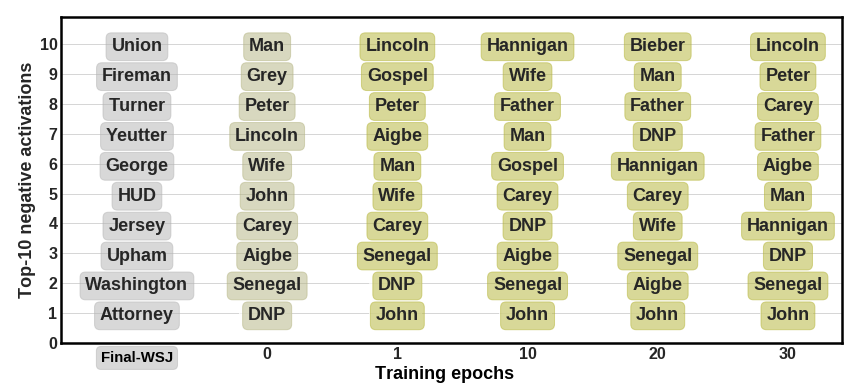}}
\vspace{-0.3cm}
\caption{Top-10 words activating unit-169 of standard fine-tuning scheme, before fine-tuning (Final-WSJ) and during fine-tuning on TweeBank. 
}
\label{neuron_pret_top_act}
\vspace{-0.4cm}
\end{figure}

Additionally, we visualise in Fig.~\ref{neuron_pret_top_act} a concrete example of a biased neuron when transferring from newswire to Tweets domain. 
Specifically, we show the top-10 words activating unit-169 of $\mathbf{\Phi}$ (from the standard fine-tuning baseline), \textit{before} fine-tuning (at this stage, the model is trained on the source-dataset WSJ) and \textit{during fine-tuning} on the TweeBank dataset. 
We can observe that this unit is highly sensitive to proper nouns (\textit{e.g.}, \textit{George} and \textit{Washington}) before fine-tuning, and to words with capitalised first-letter whether the word is a proper noun or not (\textit{e.g.}, \textit{Man} and \textit{Father}) during fine-tuning on TweeBank dataset. 
Indeed, we found that most of tokens with upper-cased first letter are mistakenly predicted as proper nouns  (\textit{PROPN}) in the standard fine-tuning scheme. 
In fact, in standard English, inside sentences, only proper nouns start with upper-cased letter thus fine-tuning the pre-trained model fails to slough this pattern which is not always respected in Tweets.\\

\begin{figure}[tb!]
\centerline{\includegraphics[scale=0.26]{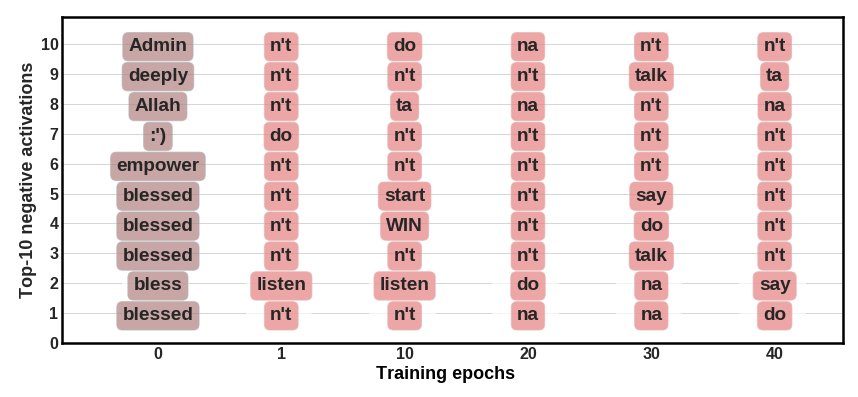}}
\vspace{-0.3cm}
\caption{Top-10 words activating unit-99 of the random branch of PretRand, before and during training.}
\label{neuron_random_top_act}
\vspace{-0.5cm}
\end{figure}

\noindent
\textbf{Unique units emerge in random branch}\\
\noindent
Finally, we highlight the ability of randomly initialised units to learn patterns specific to the target-dataset and not learned by the pre-trained ones because of their bias problem. 
To do so, we visualise unique units -- \textit{i.e.}, random units having a correlation lower than 0.4 with pre-trained ones -- emerging in the random branch. 
While only one shown in Fig.~\ref{neuron_random_top_act},  many unique units have been learned by the random branch of our PretRand model: 37.5\% 
of the 200 random units have correlation lower than 0.4 with the pre-trained ones. 
Regarding unit-99, it is highly discriminative to tokens \textit{"na"}, \textit{"ta"} and \textit{"n't"}. 
Indeed, in TweeBank, words like \textit{"gonna"} (going to) are tokenized into two tokens: \textit{"gon"} and \textit{"na"}, with the later annotated as a \textit{particle} and the former as a \textit{verb}. 
Importantly, not even one unit from the standard fine-tuning scheme has been found firing on the same important and target-dataset specific pattern.

\section{Conclusion}
\label{sec:conclusion}
\vspace{-0.3cm}

In this paper, we introduced a method to improve fine-tuning using 3 main ideas: adding random units and jointly learn them with pre-trained ones; normalising the activations of both to balance their different behaviours; applying learnable weights on both predictors to let the network learn which of random or pre-trained one is better for every class. 
We have demonstrated its effectiveness on domain adaptation from newswire domain to three commonly used Tweets-datasets for POS tagging.  



\bibliography{naaclhlt2019}
\bibliographystyle{acl_natbib}

\end{document}